\documentclass[10pt,twocolumn,letterpaper]{article}
\usepackage{cvpr}               
\usepackage{graphicx}
\usepackage{amsmath}
\usepackage{amssymb}
\usepackage{booktabs}
\usepackage{times}
\usepackage{epsfig}
\usepackage{graphicx}
\usepackage{amsmath}
\usepackage{amssymb}
\usepackage{multirow}
\usepackage{bm}
\usepackage{array, boldline, makecell, booktabs}
\usepackage[pagebackref,breaklinks,colorlinks]{hyperref}

\usepackage[capitalize]{cleveref}
\crefname{section}{Sec.}{Secs.}
\Crefname{section}{Section}{Sections}
\Crefname{table}{Table}{Tables}
\crefname{table}{Tab.}{Tabs.}

\def\confName{CVPR}
\def\confYear{2022}

\DeclareUnicodeCharacter{2212}{-}

\newcommand{\bc}{\bm{c}}
\newcommand{\z}{\bm{z}}
\newcommand{\p}{\bm{p}}

\newcommand{\bP}{\mathbf{P}}

\newcommand{\bF}{\mathbf{F}}

\newcommand{\R}{\mathbb{R}}

\newcommand{\cons}{\mathrm{Cons}}
\newcommand{\cR}{\mathcal{R}}
\newcommand{\cB}{\mathcal{B}}
\newcommand{\cU}{\mathcal{U}}

\newcommand{\cL}{\mathcal{L}}
\newcommand{\cT}{\mathcal{T}}

\newcommand{\ptd}{PointDisc}

\begin{document}

\title{Point Discriminative Learning for Data-efficient 3D Point Cloud Analysis}

\author{Fayao Liu$^1$ \;\; Guosheng Lin$^2$ \;\; Chuan-Sheng Foo$^{1,3}$ \;\; Chaitanya K. Joshi$^4$\thanks{Work done while at Institute for Infocomm Research.} \;\; Jie Lin$^1$\\
$^1$Institute for Infocomm Research, A*STAR \;\;\;
$^2$Nanyang Technological University  \\
$^3$Centre for Frontier AI Research, A*STAR  \;\;\;  
$^4$University of Cambridge\\
}

\maketitle

\begin{abstract}
3D point cloud analysis has drawn a lot of research attention due to its wide applications.
However, collecting massive labelled 3D point cloud data is both time-consuming and labor-intensive.
This calls for data-efficient learning methods.
In this work we propose \textbf{\ptd}, a point discriminative learning method to leverage self-supervisions for data-efficient 3D point cloud classification and segmentation.
\ptd~imposes a novel point discrimination loss on the middle and global level features produced by the backbone network.
This point discrimination loss enforces learned features to be consistent with points belonging to the corresponding local shape region and inconsistent with randomly sampled noisy points. 
We conduct extensive experiments on 3D object classification, 3D semantic and part segmentation, showing the benefits of \ptd~for data-efficient learning.
Detailed analysis demonstrate that \ptd~learns unsupervised features that well capture local and global geometry.
\end{abstract}

\section{Introduction}
Nowadays, due to the increasing demand of 3D applications such as augmented reality (AR), robotic vision, autonomous driving \etc, there is a growing surge in 3D related research.
Among the various 3D representation methods such as voxels, meshes, implicit functions \etc, point clouds have become an increasingly popular option. 
Point cloud analysis has therefore become an important research area. 
Learning discriminative and transferrable shape representations is a core problem in most point cloud analysis tasks.
Supervised methods rely on massive labelled data, which requires high annotation costs.
This calls for data-efficient learning approaches.
Towards this goal, methods such as self-supervised \cite{selfsup19,pointcontrast20,ACD20,occnet21}, semi-supervised \cite{semi21,semi21_aaai}, weakly-supervised learning \cite{Wei20,Xu20} are proposed.
We here focus on the self-supervised or unsupervised learning direction, which aims to exploit self-supervisions to reduce overreliance on massive labelled data.

{
\begin{figure} [t]  \center
\resizebox{.75\linewidth}{!} {
	\includegraphics[width=0.2\textwidth]{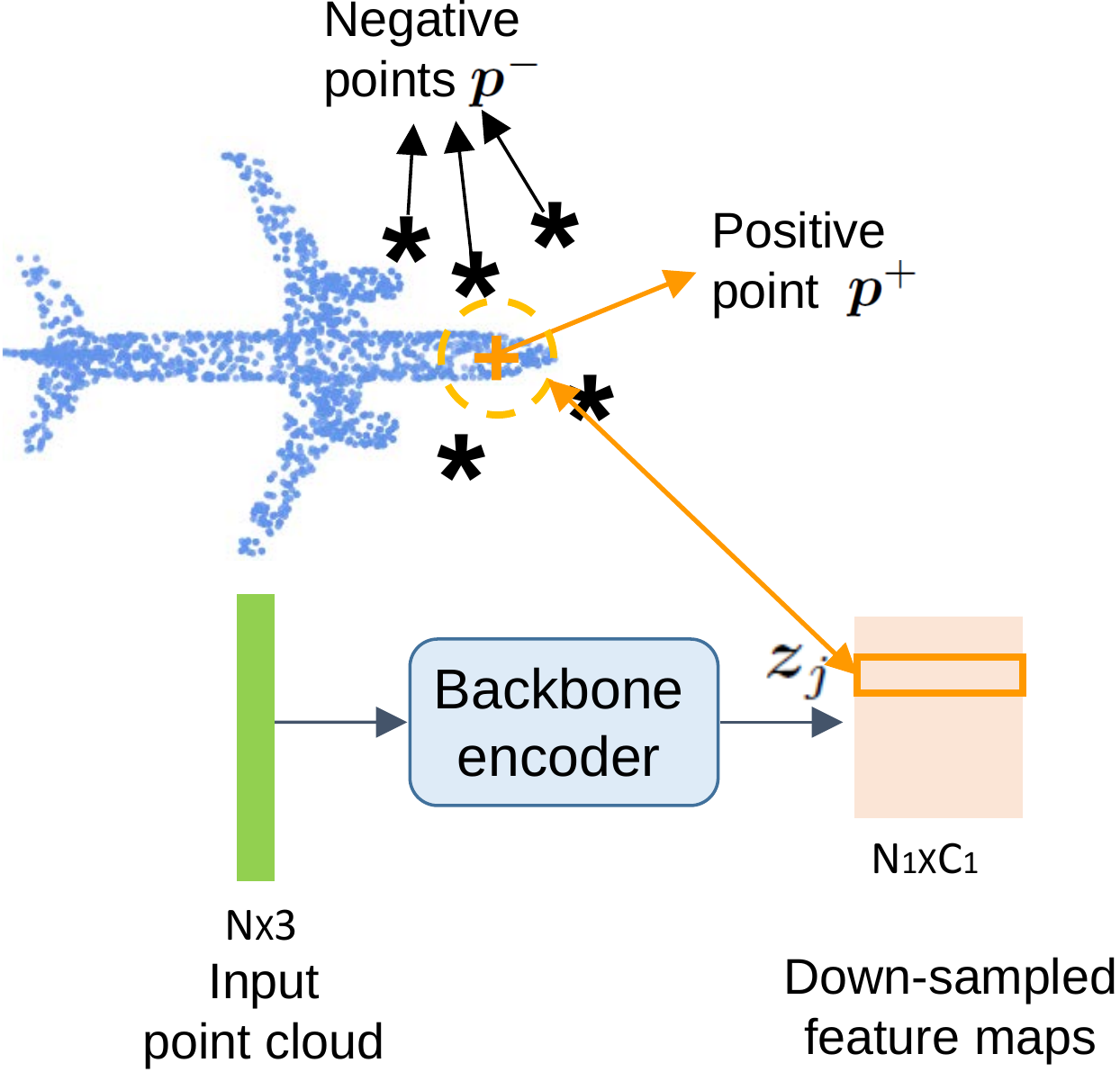}
}
\caption{An illustration of \ptd.
An input point cloud is fed to a backbone encoder to get the downsampled feature maps.
$\z_j$ is a single feature in the feature maps, which corresponds to a local region (receptive field) in the input point cloud (indicated by the dashed circle).
We define input points within this local region as positive points of $\z_j$. 
The orange plus marker shows one positive point.
We define negative points as randomly sampled noisy points, shown by black stars.
Our point discriminative learning trains the backbone encoder in an self-supervised fashion by maximizing consistency scores of $\z_j$ with $\p^+$ while minimizing consistency scores of $\z_j$ with $\p^-$.}
\label{fig:pointdisc}
\end{figure}

The recent tremendous success achieved by self-supervised representation learning in 2D domain \cite{mutualinfo19,moco20,simclr20} has inspired research on its 3D counterpart.
Various methods for self-supervised representation learning on 3D point clouds have been proposed \cite{3dgan16,lgan18,foldingnet18,3dcaps19,mapvae19,multitask19,selfsup19,clusternet19,Shi20,glr20,pointcontrast20,occnet21}. 
Most early approaches work by mapping an input point cloud into a global latent representation \cite{foldingnet18,sonet18,Shi20} or a latent distribution  in the variational case \cite{mapvae19,multitask19}  and then attempting to reconstruct the input.  
These auto-encoding based methods mostly lack effective exploitation of local geometry for self-supervision, which results in limited performance gains. 
Recently, research focus has been shifted towards developing various 3D pretext tasks for 3D self-supervised learning. 
However, due to the less structured characteristic of 3D data, designing such pretext tasks is not as straightforward as in the 2D domain such as predicting image patch orders \cite{jigsaw16}, predicting rotations \cite{rot18}, or colorizing images \cite{color16} \etc.
Therefore most existing methods for 3D self-supervised learning simply adapt techniques used in 2D domain to 3D domain. 
For example, JigSaw3D \cite{selfsup19} and Rotation3D \cite{rotation3d} are 3D versions of \cite{jigsaw16} and \cite{rot18} respectively. 
More recently, Wang \etal~\cite{occnet21} propose OcCo by reconstructing partially occluded point clouds as the pretext task, which can be seen as a 3D version of the context encoder method of \cite{occo2d}.

Lacking fully exploiting the specificity of 3D data, most existing methods show limited performance improvement. 
In this work, we propose a point discriminative learning method to leverage self-supervisions in the point cloud data, which is particularly designed for 3D point clouds. 
Our motivation is that the learned shape features should be consistent with points from the shape surface and inconsistent with points outside the shape surface. 
This is inspired by recent works on 3D implicit representations \cite{occnet19,Sitzmann19}.
Fig. \ref{fig:pointdisc} gives an illustration of our proposed method.
For a learned shape feature $\z$, we regard points within its corresponding input local region (belonging to the shape surface) as positive points and those randomly sampled noises as negative points.
We then design a point discrimination loss to maximize the consistency scores of $\z$ with their corresponding positive points, and at meantime minimize the consistency scores of $\z$ with negative points.
The consistency scores of features and points are modeled by a point consistency module, which implicitly represents the objects' surface.
Due to this novel design of point discriminative learning, our method is especially good at capturing 3D shape geometry. This is validated by our extensive experiments and detailed ablation studies.
Our contributions are:
\begin{itemize}
\item We propose \ptd, a point discriminative learning method to leverage point-level self-supervisions for data-efficient 3D point cloud classification and segmentation.
\ptd~works by enforcing learned features to be consistent with points belonging to the input point cloud using a point consistency module and a cross-entropy loss.

\item
We provide detailed analysis of \ptd\ and visually demonstrate that the unsupervised features learned by \ptd~ implicitly capture the 3D shape geometry of input point clouds.

\item
We conduct extensive experiments on 3D classification and segmentation tasks, showing that \ptd~ achieves new state-of-the-art results in various data-efficient evaluation settings. 
\end{itemize}

\section{Related work}
\paragraph{\textbf{Self-supervised representation learning on point clouds}}
Current work can be roughly classified into three categories, \ie, self-reconstruction or auto-encoding based, generative model (\eg~GAN \cite{gan}) based and self-supervised methods relying on pretext tasks beyond self-reconstruction.
Most early methods \cite{lgan18,sonet18,foldingnet18,mapvae19,multitask19,3dcaps19,glr20} belong to the first category.
These approaches mostly lack effective exploitation of local geometry supervisions, as discussed in \cite{mapvae19}.
The authors of \cite{mapvae19} then propose to capture the local geometry by multi-angle half-to-half prediction. 
Another line of 3D self-supervised representation learning methods resort to generative models like generative adversarial networks \cite{gan} as proposed in \cite{3dgan16,lgan18}. 
Approaches in the third category rely on pretext tasks beyond self-reconstruction \cite{selfsup19,rotation3d,Shi20}, which are attracting more research attention.
In \cite{selfsup19}, the authors propose a pretext task for self-supervision by re-arranging randomly shuffled 3D parts.
Poursaeed \etal~\cite{rotation3d} propose to predict the rotation angle as the pretext task.
Shi \etal~\cite{Shi20} propose a maximum likelihood
estimation method to restore the input point cloud from the one perturbed by Gaussian random noises.
Gadelha \etal~\cite{ACD20} proposes an Approximate Convex Decomposition as self-supervisions based on the observation that convexity provides cues to capture structural components.
In \cite{occnet21}, Wang \etal~propose OcCo, which performs self-supervised representation learning through completing partial point clouds that are constructed by occluding from different view-points. 
Huang \etal \cite{Huang21} propose to learn invariant representations from  self-generated or readily available temporal point cloud data  by leveraging spatial data augmentation in a self-supervised fashion.

Different from the these approaches, our method performs a point discriminative learning task to leverage self-supervisions at the point level. 
By imposing a novel point discrimination loss on different levels of the feature maps produced by the encoder, we explicitly enforce the learned features to be consistent with the global and local geometry of input point clouds. 
A recent method proposed in \cite{ParAE21} learns a maximum likelihood network for probabilistic geometric spatial partition assignments, which also explicitly leverage the geometric nature of point cloud data like our method.
In terms of techniques, our method is completely different from \cite{ParAE21} in that we perform a novel point discrimination task while \cite{ParAE21} learns discrete generative models as the pretext task.

\paragraph{\textbf{Contrastive learning}}
Contrastive learning \cite{contrastive06} has become a powerful approach for self-supervised feature learning in the 2D domain \cite{discrim14,mutualinfo19,simclr20,moco20}.
They mainly work by relying on a designed pretext task and then perform dictionary look-up, where a query is enforced to be similar to its positive match and dissimilar to others by optimizing a contrastive loss \cite{moco20}.
Such pretext tasks include exploiting agreement between examples generated by data augmentation strategies \cite{discrim14,simclr20}, maximizing the mutual information between local patches and global images \cite{mutualinfo19} \etc.
A popular contrastive loss is the InfoNCE \cite{infonce18}, which is formulated as a classification problem and implemented with a cross-entropy loss.
Our point discrimination loss is also designed as a cross-entropy loss similar to InfoNCE, which is used for maximizing the consistency scores of learned features with positive points in a comparative manner.

In the 3D domain, the authors in \cite{clusternet19} propose an object part contrasting method termed as ClusterNet based on graph convolutional neural networks \cite{gcn17} for self-supervised feature learning.
Their method is trained to determine whether two randomly sampled parts are from the same object, which can be seen as  part-level contrastive learning.
A point-level contrastive learning method for self-supervised point cloud feature learning using the InfoNCE loss was proposed in \cite{pointcontrast20} and referred as PointContrast.
They exploit point cloud correspondences between different views generated by data augmentation techniques.
More recently, DepthContrast \cite{depthcontrast21} was proposed to pre-train on single-view depth scans by contrasting at the instance level.
Similar to these approaches,  our point discriminative learning method exploits self-supervisions in a contrastive way.
Different from them, we directly enforce the learned features to capture the local and global geometry at the point level. It does not rely on instance-level 3D data augmentation techniques but a noisy point sampling strategy.

\section{Proposed method} \label{sec:method}
Consider M point clouds $\{ \bP_i\}_{i=1,\ldots,M}$, where each $\bP_i \in \R^{N \times 3}$ with each row being a single point $\p_{k} \in \R^3$.
Here $N$ is the total number of points in a single point cloud.
We aim to design self-supervised representation learning methods for data-efficient point clouds analysis.
Towards this goal, we propose a point discriminative learning method by 
maximizing the consistency scores of learned features with positive points belonging to the 3D shape. Meanwhile we minimize the consistency scores of features with randomly sampled negative points.
We expect this discrimination task to provide supervisions for learning representations that well capture global and local shapes.

\begin{figure} [t]  \center
\resizebox{1\linewidth}{!} {
\begin{tabular}{c}
	\includegraphics[width=1\textwidth,height=.75\textwidth]{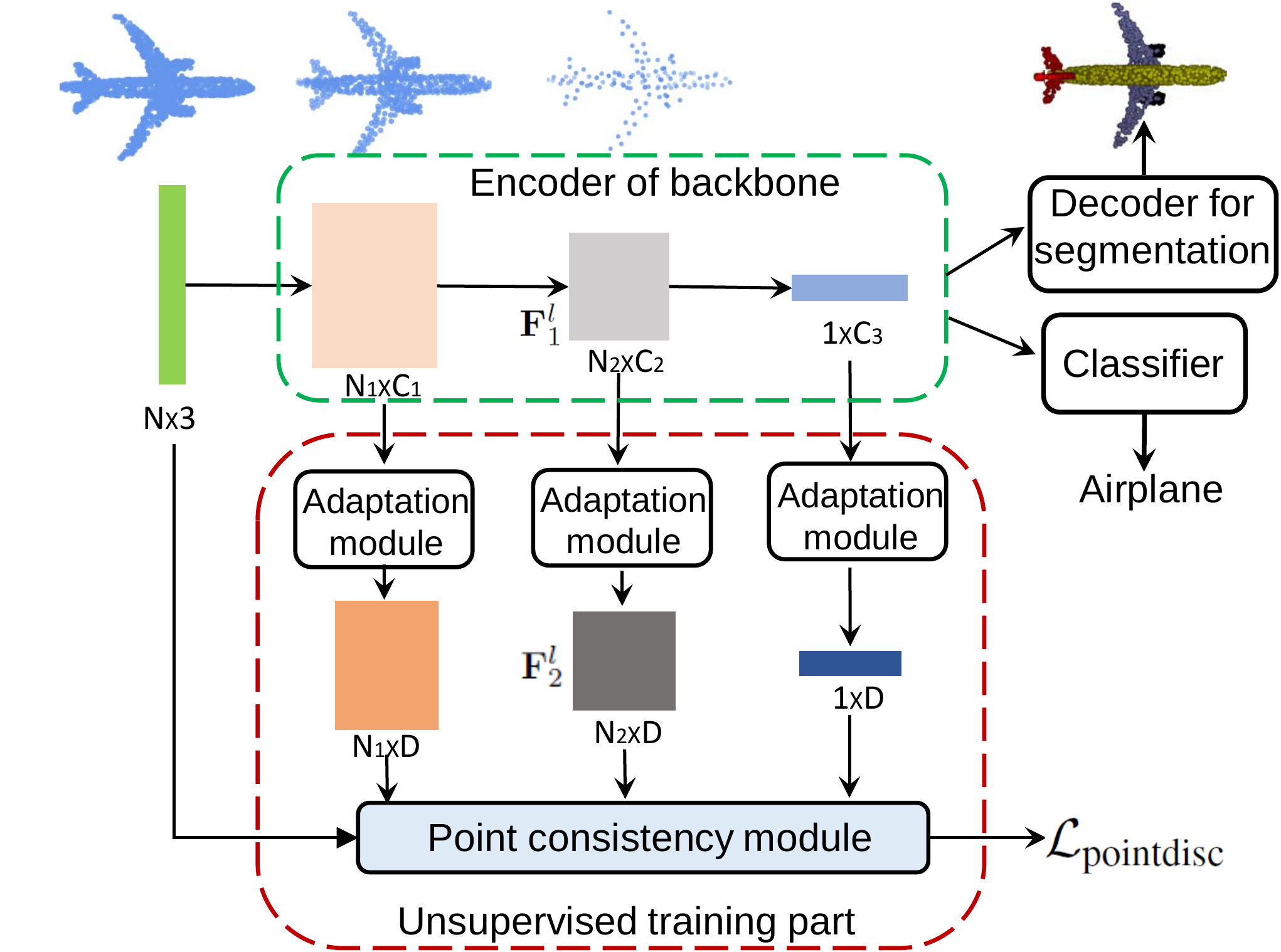}
\end{tabular}
} 
\caption{An overview of \ptd.
Our method adds an additional unsupervised training part (red box) to the backbone encoder (green box), which enforces a point discrimination loss on the middle and global level features output by the backbone encoder.
This point discrimination loss maximizes the consistency scores of learned features with corresponding points belonging to the local shape. Meanwhile it minimizes consistency scores of learned features with randomly sampled noisy points.
Once trained, the unsupervised training part can be discarded during supervised training of the classifier or decoder. } \label{fig:model}
\end{figure}

\subsection{Method overview} \label{sec:overview}
Our method is designed on top of the popular PointNet2 backbone \cite{pointnet2},
with an overview shown in Fig. \ref{fig:model}.
The encoder of the backbone is denoted by the green box while our unsupervised training part is outlined by the red box. 
The input point cloud is first encoded into a sequence of down-sampled feature maps and then decoded into a category label for classification or point-wise predictions for segmentation.

We propose a point discriminative learning method to train the backbone encoder in a self-supervised way.
Specifically, for feature maps produced by the $l$-th layer of the backbone encoder $\bF_1^l \in \R^{N_l \times C_l}$, we first use an adaptation module to map $\bF_1^l$ to a unified dimension $D$ to obtain new feature maps $\bF_2^l \in \R^{N_l \times D}$.
We then define a point discrimination loss over the features $\z_j \in \R^D$ in $\bF_2^l$ by discriminating positive points $\p^+$  from negative points $\p^-$ conditioning on $\z_j$. $\p^+, \p^- \in \R^3$ are 3 dimensional point coordinates.
It is implemented with a point consistency module and a cross-entropy loss.
The positive points $\p^+$ are defined as points that are within the corresponding local region (receptive field) of $\z_j$ in the input point cloud. Negative points $\p^-$ are generated by adding random noises to positive points (We provide discussions on whether noisy points sampled near the shape surface harm point discriminative learning in Sec. \ref{sec:ablation}).
The proposed self-supervised point discriminative learning enforces learned feature descriptors to be consistent with the local and global geometry of the input point cloud.  
After the per-training stage, the unsupervised training part can be discarded. We can use the learned encoder as an initialization for supervised training of the whole backbone to perform downstream tasks including point cloud classification and segmentation.

\subsection{Point discriminative learning} \label{sec:details}
We propose a point discriminative learning method for self-supervised point cloud representation learning. As shown in Fig. \ref{fig:model}, this self-supervised loss can be imposed on the middle level and global level point feature maps output by the encoder network. 
When imposed on the global feature vector, it enforces the learned global feature to capture the overall shape of the input point cloud.
We present details of each component of our method in the following.


\paragraph{\textbf{Point discrimination loss}}
We define the point discrimination loss for a training mini-batch as:
\begin{equation} \label{eq:loss_batch}
\cL_{\text{pointdisc}} = \sum_{j=1}^{|\cB|} \cL (\z_j).
\end{equation} 
Here $\z_j \in \R^D $ are feature vectors in feature maps $\bF_2^l$ with $l$ indicating the layer on which we intend to impose our point discrimination loss.
$|\cB|$ is the total number of $\z_j$ we use to calculate the loss in one mini-batch.

We define $\cL(\z_j)$ by enforcing $\z_j$ to be consistent with positive points $\p_i^+ \in \cR(\bc_j)$.
Here $\bc_j \in \R^3$ denotes the point coordinate associated with $\z_j$ in the downsampled feature map, which corresponds to a local region (receptive field) in the input point cloud.
$\cR(\bc_j)$ is the set of input points that are within a neighborhood of $\bc_j$.
To measure the agreement between features and points, we propose a point consistency module $\cons(\cdot)$ to output the consistency score.
$\cons(\cdot)$ is modelled as a neural network with details given next.
We then enforce $\cons(\z_j, \p_i^+)$ to be larger than any $\cons(\z_j, \p^-)$ with $\p^-$ being randomly sampled negative points.
To achieve this goal, we define $\cL(\z_j)$ in a discrimination fashion with the cross-entropy loss: 
\begin{align} \label{eq:loss_p}
&\cL (\z_j) = -\frac{1}{K}\sum_{i=1}^K  \log  \frac{\exp{(\cons(\z_j, \p_i^+)/\tau)}}{\exp{(\cons(\z_j, \p_i^+)/\tau)} + S^-}, \notag \\
&\text {with}\;\;  S^- = \sum_{t=1}^T \exp{(\cons(\z_j, \p_t^-)/\tau)}.
\end{align} 
Here $K$, $T$ are the numbers of positive and negative points sampled respectively. $\tau$ is the temperature hyper-parameter. During implementation, for each point cloud in a single mini-batch, we randomly sample 1000 groups of $(\z_j, \p_1^+, \ldots, \p_K^+)$ for calculating the point discrimination loss. Therefore, we have $|\cB|=1000 \times B$ in Eq.~\ref{eq:loss_batch} where $B$ is the training batch size of the input point clouds.

\paragraph{\textbf{Point consistency module}}
The point consistency module $\cons: \R^D \times \R^3 \rightarrow \R$ outputs the consistency score of a learned feature $\z$ with a point $\p$. 
We use a neural network to achieve this purpose. The network architecture is shown in Fig. \ref{fig:cbn}.
The concatenation $\hat{\z}=[\p, \z]$ is used as the network input.
Following the architecture used in \cite{occnet19,shapegf20}, we first map the input $\hat{\z}$ to a hidden dimension of 256 using a fully connected layer.
It is then followed with a pre-activation ResNet block \cite{resnet} with conditional batch normalization (CBN) \cite{cbn17}.  
Specifically, the ResNet block consists of two sets of CBN, a ReLU activation layer and a fully-connected (FC) layer with dimension of 256 for the hidden layer.
The output of the ResNet block is fed to another set of CBN, ReLU and FC layer to produce the 1-dimensional consistency score, which is used for calculating the point discrimination loss in Eq. \ref{eq:loss_p}.

For the CBN module, it takes $\hat{\z}$ as the condition code, which is passed through two FC layers to output the batch normalization parameters, \ie, 256-dimensional vectors $\gamma(\hat{\z})$ and $\beta(\hat{\z})$.
Then for an input $x$, the output of CBN is calculated as: 
\begin{equation} \label{eq:cbn}
\text{CBN}(x) = \gamma(\hat{\z}) \frac{x - \mu}{\sigma} + \beta(\hat{\z}),
\end{equation}
where $\mu$ and $\sigma$ are the mean and standard deviation of the batch data.  
Compared with standard BN where $\gamma$ and $\beta$ are fixed after learning, here CBN produces dynamic $\gamma$ and $\beta$ for different inputs using a neural network.
Our experiments show that adding the CBN module leads to more stable training and better generalization. 
We show this point consistency module design outperforms the one without CBN in the ablation studies in Sec. \ref{sec:ablation}.

 {\small
\begin{figure} [t]  \center
\resizebox{.75\linewidth}{!} {
	\includegraphics[width=0.4\textwidth]{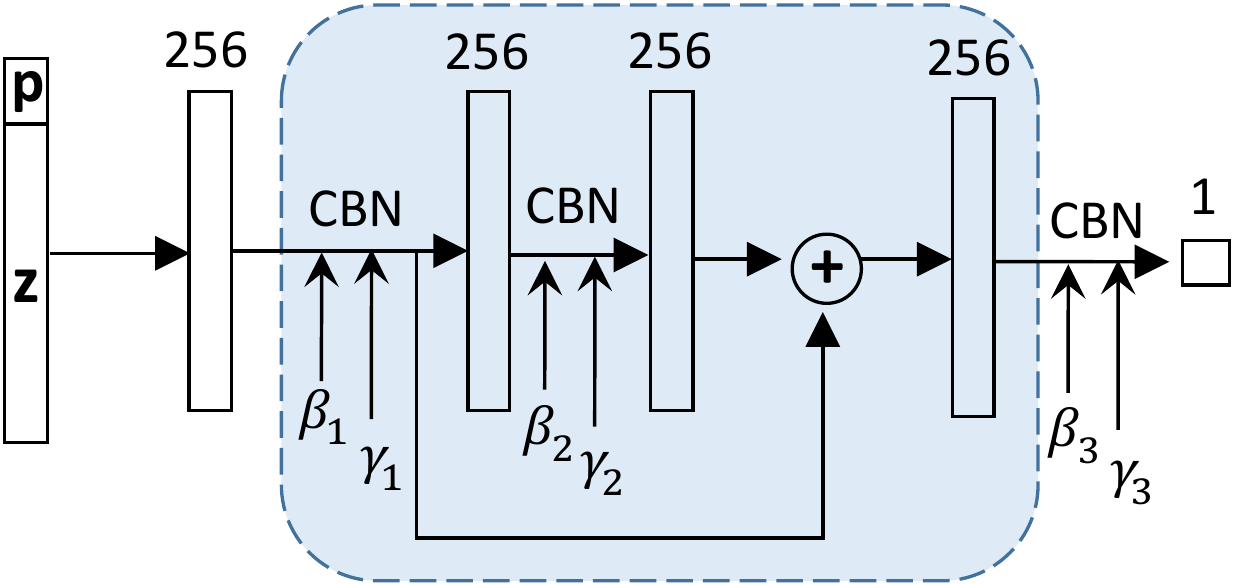}
}
\caption{Network architecture of our point consistency module $\cons(\cdot)$. The blue box indicates a ResNet module. CBN is the conditional batch normalization \cite{cbn17}. }
\label{fig:cbn}
\end{figure}
}

\paragraph{\textbf{Positive and negative point sets construction}}
For a particular feature $\z_j$ in the feature maps $\bF_2^l$,  we define its positive points $\p_i^+$ as the input points that are within a neighborhood region of $\bc_j$. 
Here $\bc_j$ is the point coordinate corresponding to $\z_j$, which is one of the output by the set abstraction layer in PointNet2. 
We then conduct ball query with a predefined radius of $\bc_j$ to find $\cR(\bc_j)$.
The negative point set for $\z_j$ is denoted as $\cT(\cR(\z_j))$ with $\cT(\cdot)$ being a random perturbation operation. 
We define $\cT(\cdot)$  as adding some random noise $\epsilon \sim \cU[-a, a]$ to the input coordinates. 
$\cU$ denotes a uniform distribution with $a$ being a scalar parameter.  
Gaussian noise can be an alternative option here.
We conduct extensive ablation studies on the negative point set construction strategies in Sec.~\ref{sec:ablation}.

\section{Experiments}

\subsection{Experimental setup}
We evaluate \ptd~on 3D object classification, semantic segmentation and part segmentation tasks. 
Our baseline is PointNet2 with random initializations. 

\noindent{\textbf{Datasets}} We conduct 3D object classification on ModelNet \cite{modelnet}, ScanObjectNN \cite{scanobjnn} and ScanNet10 \cite{scannet10}.
ModelNet40 and ModelNet10 are composed of CAD models, with each containing 9832/2468 and 3991/908 training/test objects coming from 40 and 10 classes respectively.
ScanObjectNN and ScanNet10 are real world datasets containing point cloud scans with occlusions and noises.
For semantic segmentation, we use the S3DIS benchmark \cite{s3dis}, which consists of point cloud scans from 6 areas covering 271 rooms and 13 semantic classes.
For part segmentation, we evaluate on ShapeNetPart \cite{shapenetpart}, which consists of 16881 objects from 16 categories, with each object segmented into 2 to 6 parts. There are 50 parts in total.
We follow the standard train/test/val splits for all tasks unless otherwise stated.
For evaluation metrics, we  use the global accuracy (Acc) for classification and mean Intersection-over-Union (mIoU) for segmentation tasks.

\noindent{\textbf{Implementation details}}
The adaptation module is designed as a 2-layer multi-layer perceptron (MLP) network with batch normalization and ReLU activation layers.
Each layer has a dimension of 256 and the output dimension $D$ is set to 256.
The outputs of the adaptation module are $L_2$ normalized.
For finding $\cR(\bc_j)$, we use ball query within a radius which equals to the radius parameter of the grouping layer in PointNet2 \cite{pointnet2}.
We normalize the input point clouds by centering their bounding boxes to the origin and scaling them to ensure that all points range within the cube $[-1, 1]^3$ following \cite{shapegf20}. The uniform sampling parameter $a$ for negative points construction is set to 1. 
We train the model with an Adam optimizer and a batch size of 24. 
The learning rates for self-supervised pre-training and fine-tuning  are  initialized as 0.001 and 0.0005 respectively and scheduled with exponential decay.
The temperature parameter $\tau$ in Eq. \ref{eq:loss_p} is set to 0.1.

\begin{table}[!htb] 
\centering
\resizebox{1\linewidth}{!} {
\begin{tabular}{l|c|c|c}
\hlineB{2.5}
Method &ModelNet40 &ScanNet  &ScanObjectNN   \\
\hline
\hline
DGCNN \cite{occnet21}  &92.5$\pm$0.4  &76.1$\pm$0.7  &82.4$\pm$0.4 \\
JigSaw3D \cite{selfsup19} &92.3$\pm$0.3 &77.8$\pm$0.5  &82.7$\pm$0.8   \\
OcCo \cite{occnet21}  &\textbf{93.0$\pm$0.2}  &78.5$\pm$0.3 &83.9$\pm$0.4  \\
\hline
Baseline &91.2$\pm$0.3 &77.5$\pm$0.4  &82.1$\pm$0.3  \\
\textbf{\ptd~(Ours)}    &92.6$\pm$0.2  &\textbf{80.1$\pm$0.3 }   &\textbf{86.2$\pm$0.3} \\
\hlineB{2.5}
\end{tabular}
}
\caption{3D object classification (Acc) on three datasets by pre-training on ModelNet40.  The results are reported as mean$\pm$ste (standard error) over 3 runs.
Our method achieves the best results on ScanNet and ScanObjectNN.
}  \label{tab:trans}
\end{table}

\begin{table}
\centering
\resizebox{.86\linewidth}{!} {
\begin{tabular}{l|c|c}
\hlineB{2.5}
Method  &ModelNet40 &ModelNet10 \\
\hline\hline
T-L Network \cite{tlnet16}  &74.40 &- \\
3DGAN \cite{3dgan16}  &83.30 &91.00 \\
VSL \cite{vsl18}  &84.50  &91.00 \\
VIPGAN \cite{vipgan19}  &91.98 & 94.05 \\
MRTNet \cite{mrtnet18}  &86.40 &- \\
LGAN$^{\dagger}$ \cite{lgan18}   &87.27 &92.18 \\
LGAN \cite{lgan18}   &85.70 &95.30 \\
PointCapsNet \cite{3dcaps19}  &88.90 &-\\
FoldingNet$^{\dagger}$ \cite{foldingnet18}  &88.40 &94.40 \\
FoldingNet \cite{foldingnet18}   &84.36 &91.85 \\
ClusterNet \cite{clusternet19} &86.80 &93.80 \\
Multi-task$^{\dagger}$ \cite{multitask19} &89.10 &- \\
MAP-VAE$^{\dagger}$ \cite{mapvae19}  &90.15 &94.82 \\
Rotation3D \cite{rotation3d}  &91.84 &-\\
JigSaw3D$^{\dagger}$ \cite{selfsup19} &90.69  &94.52 \\
GLR\footnotemark[1] \cite{glr20}  &92.22 &94.82 \\
ACD$^{\dagger}$ \cite{ACD20} &89.80 &- \\
OcCo \cite{occnet21}  &89.20  &- \\
STLR$^{\dagger}$ \cite{Huang21} &90.90 &- \\
ParAE$^{\dagger}$ \cite{ParAE21}  &91.60 &- \\ 
\hline
\textbf{\ptd\ (Ours)}  &\textbf{92.30}  &\textbf{95.37} \\
\hlineB{2.5}
\end{tabular}
}
\caption{Linear SVM results on ModelNet40 and ModelNet10. Our \ptd~outperforms all methods. ${\dagger}$ indicates the method is trained on ShapeNet.}  \label{tab:lsvm}
\end{table}

\subsection{3D object classification}
\paragraph{\textbf{Comparisons with previous work} }
We first perform object classification on  ModelNet40, ScanObjectNN and ScanNet10. 
We use a single model pre-trained on ModelNet40 as initialization and fine-tune it on each of the 3 datasets. The results are reported over 3 runs and shown in Table \ref{tab:trans}.
Our \ptd~performs consistently better than the baseline while outperforming JigSaw3D and OcCo on ScanNet and ScanObjectNN with considerable margins.
OcCo achieves the best results on ModelNet40. We note that in terms of relative accuracy boost over the baseline, our method brings an accuracy improvement of 1.4 compared to 0.5 by OcCo. 
It is worth noting that our \ptd~performs quite well in the  cross-dataset  setting, \ie, from ModelNet40 to ScanNet and ScanObjectNN.
Since ScanNet and ScanObjectNN are real world point cloud datasets with occlusions and noises, this experiment demonstrate that pre-training on CAD models (ModelNet40) with \ptd~can help the recognition of real world datasets.

\footnotetext[1]{\footnotesize{The results of \cite{glr20} are reported as using the same backbone as us. Better results are achieved in \cite{glr20} by using 5 times wider networks.}}

We further report the results of training a linear support vector machine (SVM) \cite{svm} over the features learned by our \ptd.
Following \cite{glr20}, we aggregate features from different layers to train a linear SVM.
The compared results on ModelNet40 and Model10 are reported in Table \ref{tab:lsvm}.
As we can see, our method achieves the best results on both datasets, outperforming recent methods \cite{occnet21,ParAE21}.

\begin{figure}[!t]  \center
\resizebox{1\linewidth}{!} {
\begin{tabular}{cc}
\includegraphics[width=0.5\linewidth,height=0.36\linewidth]{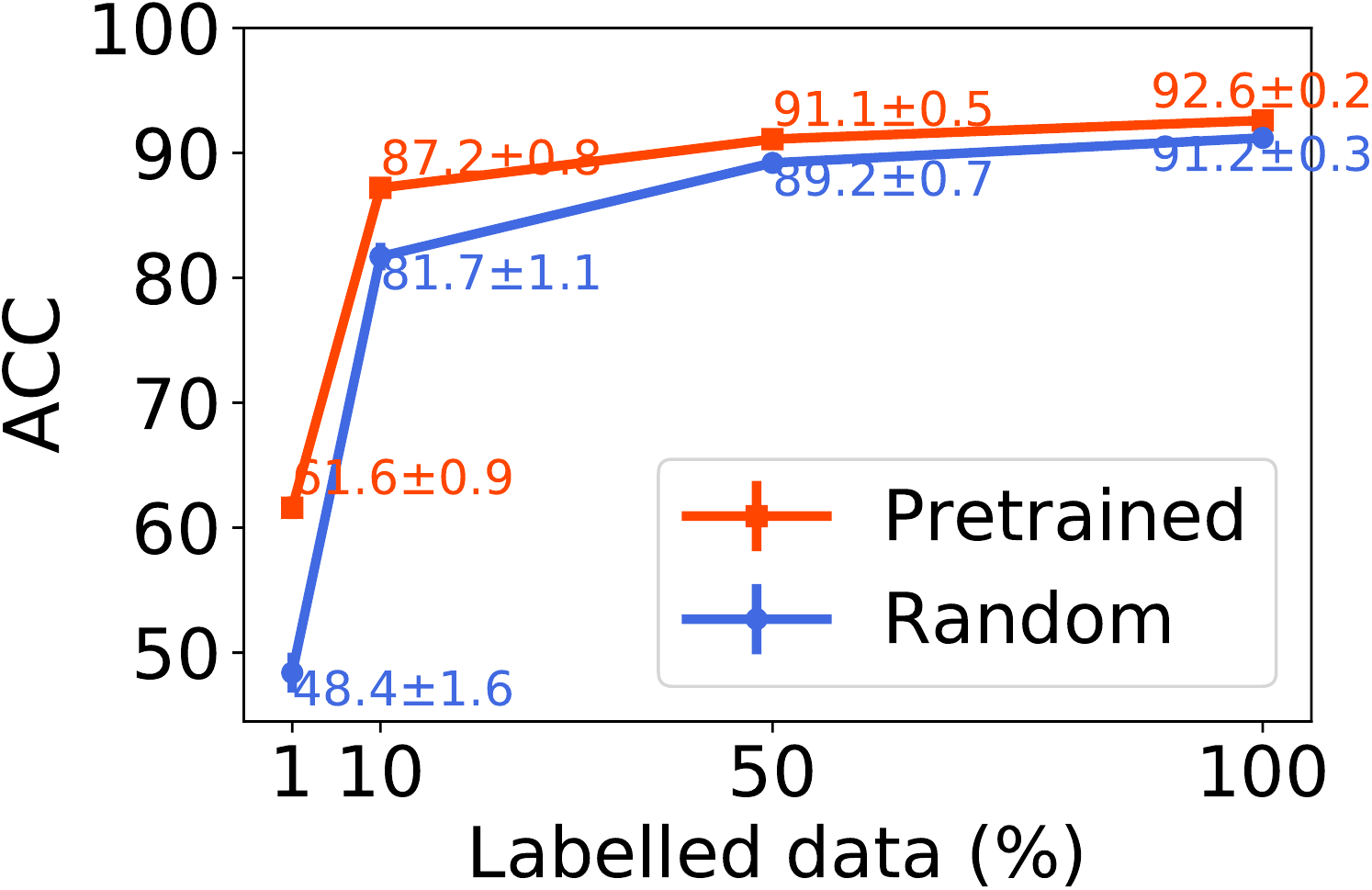}
&\includegraphics[width=0.5\linewidth,height=0.36\linewidth]{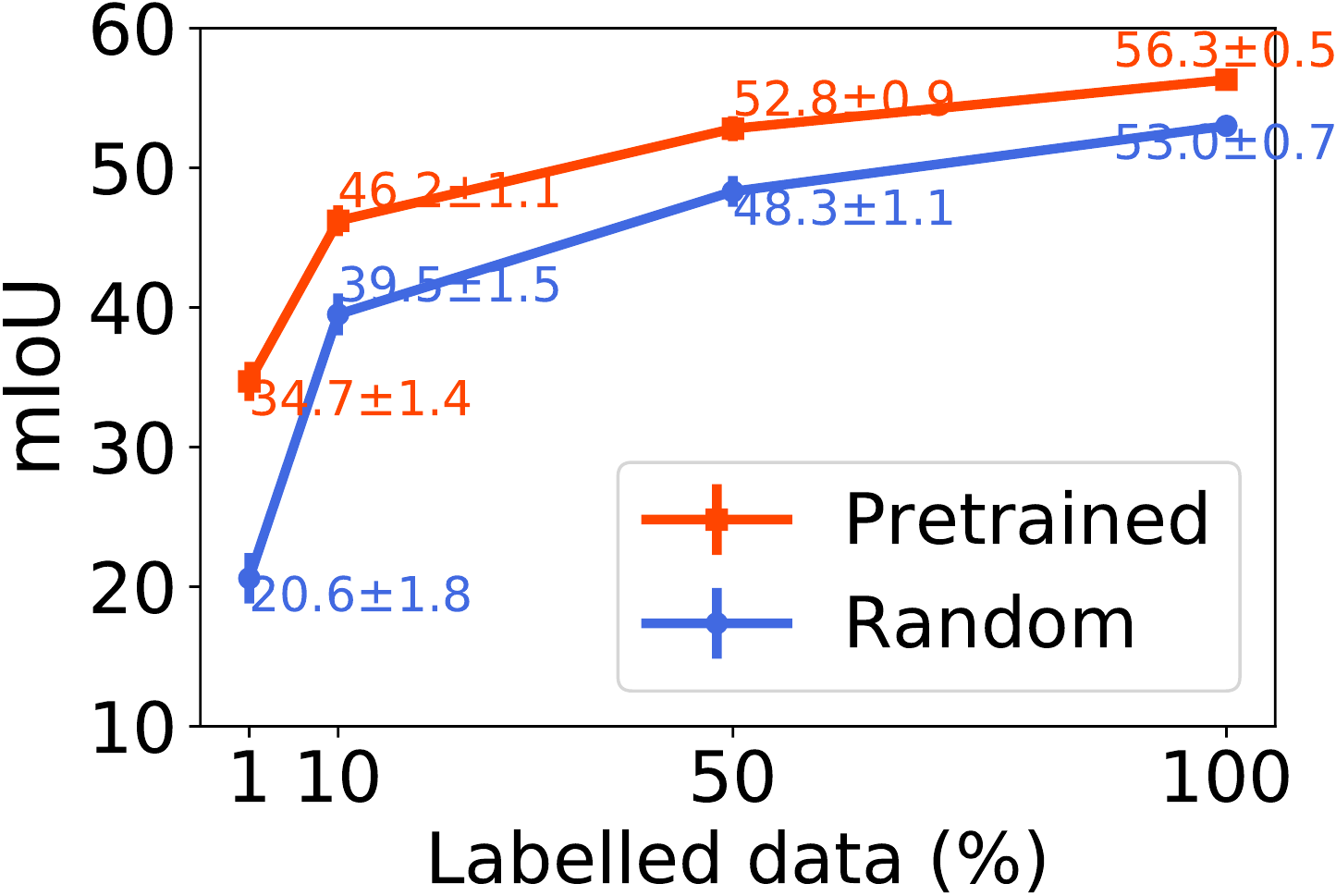}
\end{tabular} }
\caption{ Left: Classification results on ModelNet40 wrt. 1\%, 10\%, 50\%, 100\% labelled training data. Right: Segmentation results on S3DIS Area 5 wrt. 1\%, 10\%, 50\%, 100\% labelled training data. }
\label{fig:dataefficient}
\end{figure}

\paragraph {\textbf{Evaluations under limited training budget}}
We then evaluate in a data-efficient setting, \ie, using limited labelled training data budget.
Specifically, we evaluate the baseline and our \ptd~method using 1\%, 10\% and 50\% labelled training data. Our \ptd~is pre-trained on the unlabelled ModelNet40 training set.
Fig.~\ref{fig:dataefficient} (left) shows the  mean$\pm$ste of classification accuracies over 3 runs. We can see that our method brings consistent improvements over the random initialized baseline. 
The improvement is more significant in the less data setting.
\textit{It is worth noting that using only 50\% of the labelled training data, our method achieves comparable performance with the baseline using 100\% labelled data.}

{\small
\begin{figure*}  \center
\resizebox{.78\linewidth}{!} {
\begin{tabular}{cc}
	\includegraphics[width=0.25\textwidth, height=0.13\textwidth]{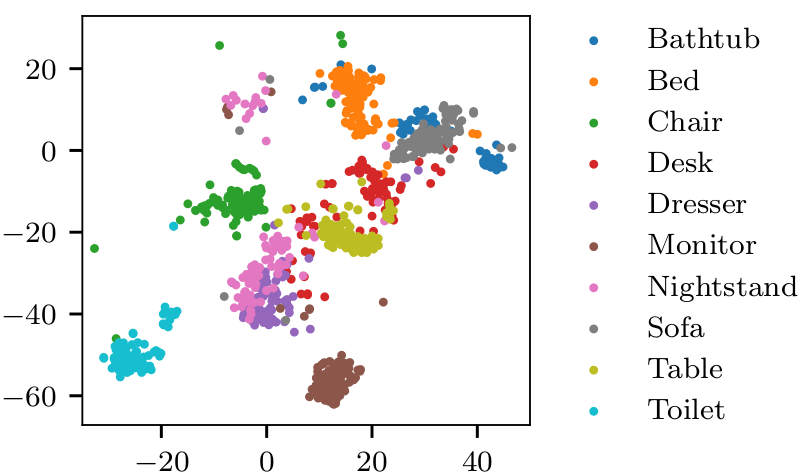}
	 &\includegraphics[width=0.24\textwidth, height=0.12\textwidth]{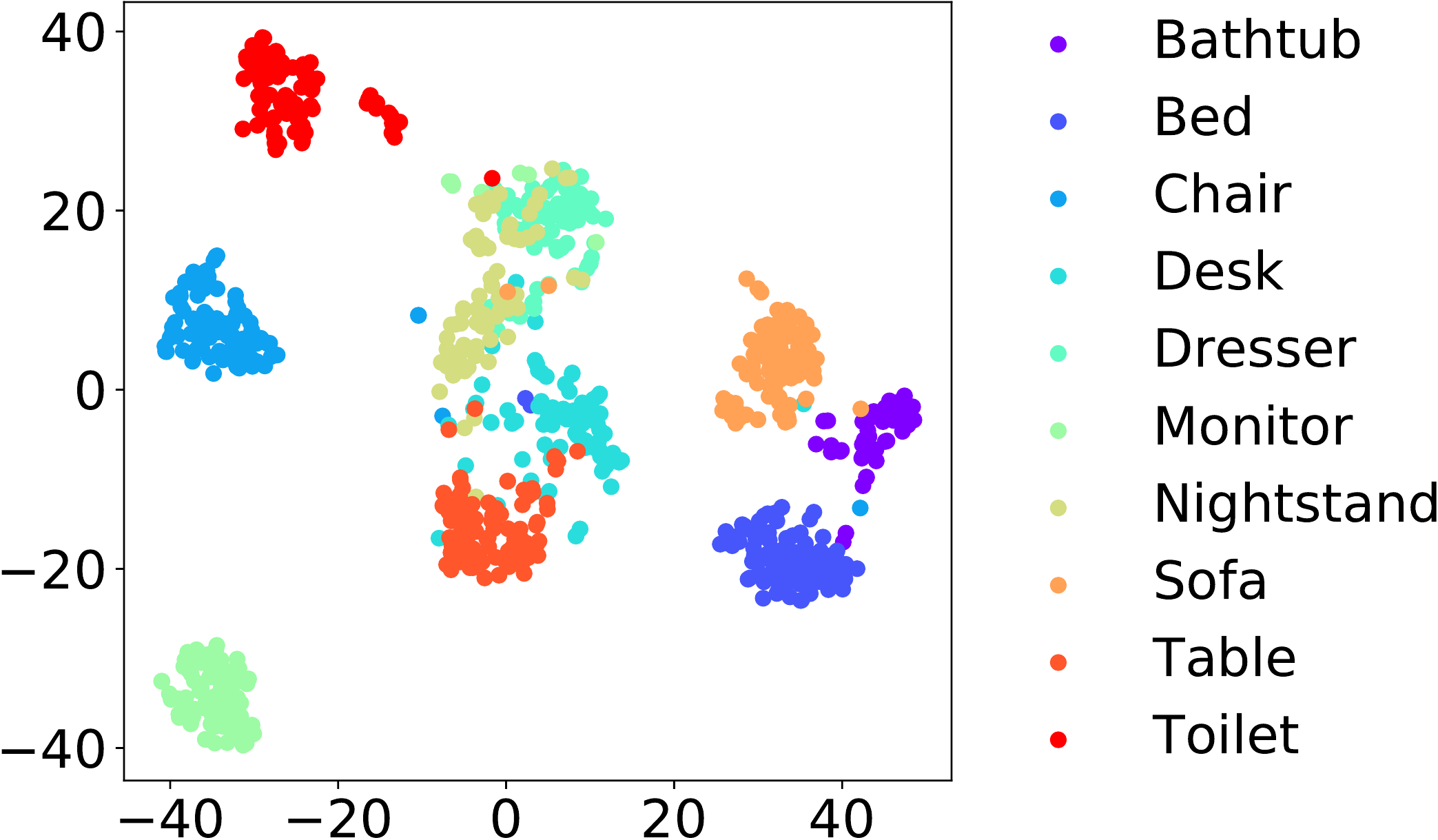} \\
\end{tabular}
}
\caption{T-SNE \cite{tsne} visualizations of learned unsupervised features on the ModelNet10 test dataset. Left: JigSaw3D \cite{selfsup19}. Right: ours. Our method produces more separable clusters for different categories.}
\label{fig:viz_tsne}
\end{figure*}
}

\paragraph{\textbf{Visualizations of learned unsupervised features}}
We visualize the learned unsupervised feature embeddings of ModelNet10 using tsne \cite{tsne} in Fig. \ref{fig:viz_tsne}. The model is trained on the ModelNet10 training set without labels and the visualization shows embeddings of the test set.
The left figure in Fig. \ref{fig:viz_tsne} shows the visualization result of JigSaw3D \cite{selfsup19} (reproduced from the original paper). The right plot shows our result. 
As we can see, for both methods, the embeddings of ``nightstand" and ``dresser" are mixed together due to their strong visual similarities. 
In general, our method produces more separable clusters than \cite{selfsup19}, which demonstrates strong feature learning capability of our point discriminative learning method.

\begin{table}
\centering
\resizebox{.63\linewidth}{!} {
\begin{tabular}{l|c}
\hlineB{2.5}
Method  &mIoU    \\
\hline
\hline
DGCNN \cite{occnet21}  &54.9$\pm$2.1    \\
JigSaw3D \cite{selfsup19}  &55.6$\pm$1.4 (0.7$\uparrow$)   \\
OcCo \cite{occnet21} &58.0$\pm$1.7 (3.1$\uparrow$) \\
\hline
Baseline    &56.1$\pm$1.9   \\
\textbf{\ptd~(Ours)}     &\textbf{60.4 $\pm$1.2 (4.3$\uparrow$)}  \\
\hlineB{2.5}
\end{tabular}
}
\caption{Semantic segmentation results (mIoU) on  S3DIS across 6 folds over 3 runs.  Our method  achieves the best performance and brings more significant mIoU boost over the baseline compared to JigSaw3D and OcCo.}  \label{tab:semseg_6fold}
\end{table}

\begin{table*} 
\centering
\resizebox{.88\linewidth}{!} {
\begin{tabular}{l|c|c|c|c|c}
\hlineB{2.5}
\multirow{2}{*}{Method} &\multicolumn{5}{c}{Supervised training area}  \\
\cline{2-6}
&Area 1  &Area 2  &Area 3  &Area 4  &Area 6  \\ 
\cline{2-6}
\hline\hline
DGCNN \cite{selfsup19} &43.6 &34.6 &39.9 &39.4 &43.9 \\ 
JigSaw3D \cite{selfsup19} &44.7 (1.1$\uparrow$) &34.9 (0.3$\uparrow$) &42.4 (2.5$\uparrow$) &39.9 (0.5$\uparrow$) &43.9 (0.0$\uparrow$) \\ 
\hline
Baseline &43.0$\pm$0.5 &33.7$\pm$0.2 &39.5$\pm$0.1  &41.0$\pm$0.4 &43.5$\pm$0.2 \\ 
\textbf{\ptd~(Ours)}    &\textbf{50.4$\pm$0.1 (6.6$\uparrow$)}  &\textbf{39.2$\pm$0.2 (5.5$\uparrow$)} &\textbf{47.7$\pm$0.2 (8.2$\uparrow$)} &\textbf{46.4$\pm$0.3 (5.4$\uparrow$)}&\textbf{48.6$\pm$0.2 (5.1$\uparrow$)} \\ 
\hlineB{2.5}
\end{tabular}
}
\caption{Semantic segmentation results (mIoU) on  S3DIS by testing on Area 5. All results are obtained by supervised training on different areas and testing on Area 5. We report our results as mean$\pm$ste (standard error) over 3 runs. Our method achieves the best performance and brings much more significant mIoU boost over the baseline compared to JigSaw3D \cite{selfsup19}.
}  \label{tab:semseg}
\end{table*}

\subsection{3D Semantic segmentation}
\paragraph{\textbf{Comparisons with previous work} }
Following OcCo \cite{occnet21}, we evaluate the 6-fold cross validation performance and report the results in Table \ref{tab:semseg_6fold}. 
Our self-supervised pre-training is conducted on the whole training set without labels. 
The results of OcCo and JigSaw3D are reproduced from \cite{occnet21}.
As we can see that our method achieves the best performance and brings more significant mIoU boost over the baseline compared to JigSaw3D and OcCo.

\paragraph{\textbf{Evaluations under limited training budget} }
We then evaluate under the limited training budget setting proposed by  JigSaw3D \cite{selfsup19}, \ie, pre-training on all areas other than Area 5 without labels, and then fine-tune the model with labelled data from a single area.
For the baseline model, we train on labelled data from different areas except Area 5 with random initialization.
Area 5 is used as the test data for all models. 
We run three times and report the mean and standard errors of mIoU scores in Table \ref{tab:semseg}.
Compared to JigSaw3D, our method achieves much better results in all settings and consistently brings notable mIoU boost over the baseline.
We further evaluate using 1\%, 10\% and 50\% of the whole training data without Area 5 and report the results in Fig.~\ref{fig:dataefficient} (right). 
We can see that our method consistently outperforms the baseline with more performance gained under less data settings.
\textit{We note that using only 50\% of the labelled training data, our method achieves comparable performance with the baseline using 100\% labelled data.}
This fully demonstrates the benefits of our method for learning with limited labelled training data budget.

\begin{table}[t]
\centering
\resizebox{.82\linewidth}{!} {
\begin{tabular}{l|c|c}
\hlineB{2.5}
Method  &1\% training &5\% training    \\
\hline
\hline
SONet \cite{sonet18} &64.0  &69.0  \\
PointCapsNet \cite{3dcaps19} &67.0  &70.0 \\
Multi-task \cite{multitask19} &68.2 &77.7  \\
PointContrast \cite{pointcontrast20}  &74.0   &79.9 \\
ACD \cite{ACD20} &75.7 &79.7 \\
\hline
\textbf{\ptd~(Ours)}   &\textbf{77.2$\pm$0.5}  &\textbf{81.3$\pm$0.4}  \\
\hlineB{2.5}
\end{tabular}
}
\caption{Part segmentation (mIoU) on ShapeNetPart using 1\% and 5\% labelled training data budgets.  Our method (reported as mean$\pm$ste over 3 runs) outperforms all compared methods in both settings. }  \label{tab:partseg}
\end{table}

\subsection{3D part segmentation}
3D part segmentation is a fine-grained point-wise classification task that requires detailed local geometry features.
Following prior works \cite{pointnet2,selfsup19}, we use the one-hot encoded category label of the object as an extra input for supervised training. During self-supervised learning stage, a random class label is given to each object.

We first pre-train the encoder with \ptd~on the whole ShapeNetPart training dataset and then fine-tune the encoder and decoder with different percentages of labelled data. 
The compared mIoU results under $1\%$ and $5\%$ training data budgets are reported in Table \ref{tab:partseg}.  
We compare against PointContrast \cite{pointcontrast20}, ACD \cite{ACD20} and other unsupervised learning methods \cite{sonet18,3dcaps19,multitask19} which have reported results under the same setting.
As we can see, our method achieves the best performance among all compared methods. 
Note that ACD uses the same PointNet2 backbone as us.
This well demonstrates the effectiveness of our self-supervised learning method for data-efficient part segmentation.

\section{Analysis and discussions} \label{sec:ablation}
In this section, we perform ablation studies for detailed analysis. 
Unless otherwise stated, all experiments are conducted by training our \ptd~on the ModelNet40 training set followed by a linear SVM for classification on ModelNet40 test set.
The hyperparameter C of SVM is chosen based on a validation set.

\paragraph{\textbf{Gaussian noise or uniform noise?}}
To perform point discriminative learning, we need to sample $\p_t^-$, $t=1,\ldots, T$ to construct the negative point set for each $\z_j$ (see Eq. \eqref{eq:loss_p}).
We perform a random perturbation on points in $\cR(\bc_j)$, where $\cR(\bc_j)$ denotes the set of positive points.
We conduct experiments with two types of noise: uniform and Gaussian.
The uniform noise is sampled from $\epsilon \sim \cU[-1, 1]$ while the Gaussian noise is sampled from a standard normal distribution.
We perform point discriminative learning with the two different noisy points sampling strategies.
Our conclusion is that the uniform noise leads to slightly better result than the Guassian noise (92.30 \vs~91.82).
For all later experiments, we use the uniform noise sampling.

\paragraph{\textbf{Do noisy points near the shape surface harm point discriminative learning?}}
During sampling of negative points, some sampled points may be very close to the local shape surface, which may cause confusion to the point discriminative learning.
To figure out this issue, we conduct an ablation study to exclude those points  
that are within a small distance (0.1) of $\cR(\bc_j)$.
Our experimental results show that there is no statistically significant difference between models with and without this point exclusion strategy.
Since the negative points are generated by adding uniform noises to positive points, those near-the-shape-surface points only consist of a small fraction of the sampled points.
We conclude that our negative point construction strategy does not bring negative effect to the point discriminative learning.

\begin{table} 
\centering
\resizebox{.8\linewidth}{!} {
\begin{tabular}{l|ccccc}
\hlineB{2.5}
T &1 &5 &10  &20  &30 \\
\hline
\hline
Acc &90.12 &91.87 &92.30 &92.20 &92.21 \\
\hlineB{2.5}
\end{tabular} }
\caption{Ablation of parameter T on ModelNet40.}\label{tab:ablation_T}
\end{table}

\paragraph{\textbf{Number of positive and negative points per $\z_j$}}
We study the effects of sampling different numbers of positive and negative points $K$, $T$ per $\z_j$ in Eq. \eqref{eq:loss_p}.
For $K$, we empirically find that setting $K$ to larger than 1 converges to similar performance as $K=1$.    
For $T$, we choose it from \{1, 5, 10, 20, 30\} to train our \ptd~model and then train a linear SVM for classification.
We show the results in Table \ref{tab:ablation_T}.
The results show that classification accuracy keeps improving as $T$ increases from 1 to 10.  
When $T$ is set to larger than 10, the performance shows no further gains. 
We set $K=1$ and $T=10$ in all our experiments.

\begin{table} 
\centering
\resizebox{.85\linewidth}{!} {
\begin{tabular}{l|c|c|c|c|c}
\hlineB{2.5}
Model &$l3$ &$l2$ &$l1$  &CBN  &Accuracy (\%) \\
\hline
\hline
A &\checkmark & & &\checkmark &90.32 \\
B &\checkmark &\checkmark &  &\checkmark &91.17 \\
C &\checkmark  &\checkmark  &\checkmark & &91.09 \\
D &\checkmark  &\checkmark  &\checkmark &\checkmark &\textbf{92.30} \\
\hlineB{2.5}
\end{tabular} }
\caption{Performance comparison of our method using different components. We evaluate the linear SVM  classification accuracy on the ModelNet40 dataset.}\label{tab:component}
\end{table}

\paragraph{\textbf{Component analysis}}
We analyze in detail the contributions of each component in our model. The ablation results are reported in Table \ref{tab:component}.
As shown in Fig. \ref{fig:model}, our self-supervised point discrimination loss can be imposed on features learned from any intermediate layers or the global level feature (last layer). 
Our first ablation factor is the contribution of imposing our point discrimination loss on different layers (model A, B, D in Table \ref{tab:component}). 
We use $l1$, $l2$, $l3$ to indicate the first, second and third layer respectively. 
Our second ablation factor is the conditional batch normalizatin (CBN) in the point consistency module, which is replaced by conventional batch normalization (model C in Table \ref{tab:component}).
As we can see that imposing the point discrimination loss on more intermediate layers consistently improves the classification performance. 
Compared to the model without CBN, our full model achieves the best performance. This validates the effectiveness of our point consistency module design.

\paragraph{\textbf{Visualizing local shapes captured by our learned unsupervised features}}
After the self-supervised training of \ptd, 
for an input point cloud, we can obtain its self-supervised features through the encoder.  For a particular $\z$ in the obtained middle-level feature maps, we can visualize the local shape captured by $\z$ through evaluating consistency scores of $\z$ with randomly sampled points.
In this way, we can probe whether the self-supervised features learned by our \ptd\ indeed capture local shapes of the input point cloud.
We visualize points with the top 100 highest consistency scores among 5000 uniformly sampled points in Fig. \ref{fig:vis_cons_part}.
The first row shows input point clouds, with each example showing a centriod (indicated by ``+") and its neighboring region (dashed circle) corresponding to a learned feature $\z$.
The second row shows ground-truth local shapes. 
They are positive points corresponding to $\z$ that are used to train our \ptd.
The third row shows local shapes reconstructed from our learned self-supervised feature $\z$.
We can see that the reconstructed local shapes are highly consistent with the ground-truth shapes.
This validates our claim that our method learns self-supervised representations which can well capture local geometry.

\begin{figure}[h]
    \centering
  \begin{subfigure}{1\linewidth}
    \includegraphics[width=1\textwidth]{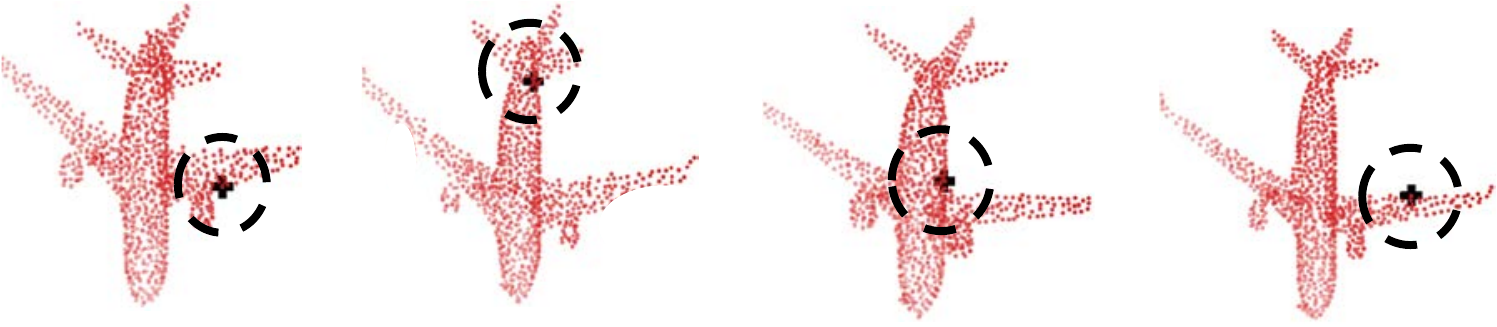} 
    \caption{Input point clouds.}
    \label{fig:vis_in}
  \end{subfigure}

    \begin{subfigure}{1\linewidth}
      \hspace{0.5 cm} \includegraphics[width=.9\textwidth]{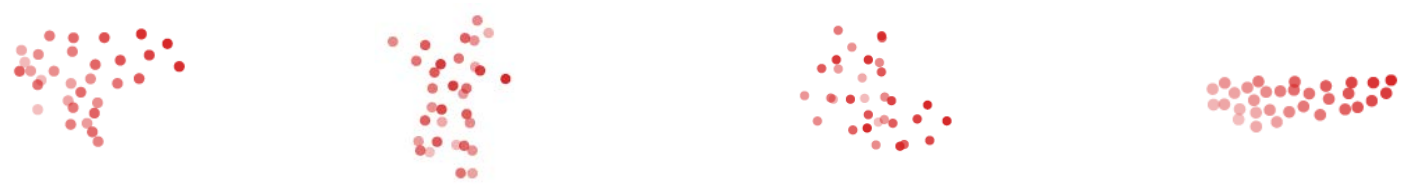} 
    \caption{Ground-truth local shapes (zoomed in).}
    \label{fig:vis_gt}
  \end{subfigure}

    \begin{subfigure}{1\linewidth}
    \hspace{0.5 cm}
    \includegraphics[width=.9\textwidth]{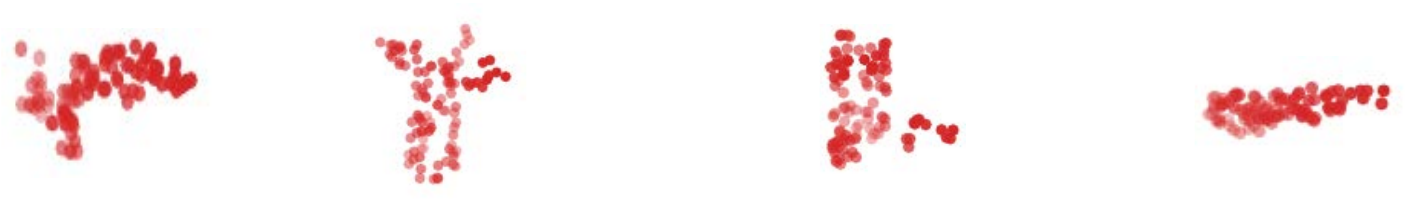} 
    \caption{Shapes reconstructed from features learned by \ptd.}
    \label{fig:vis_cons}
  \end{subfigure}
  
\caption{Visualizations of reconstructed shapes from the unsupervised features learned by  \ptd. Each column shows an input point cloud with a centroid (denoted by ``+") and its neighboring region (dashed circle) corresponding to a learned feature $\z$. }
\label{fig:vis_cons_part}
\end{figure}

\section{Conclusion}
We propose a novel point discriminative learning method for data-efficient  3D point cloud analysis.
By imposing a point discrimination loss on the middle and global level features, our method directly enforces the learned features to capture local and global shape geometry.
This is validated by our detailed ablation studies and visual demonstrations.
Extensive experiments on 3D object classification, semantic and part segmentation demonstrate that the proposed method achieves new state-of-the-art results in various data-efficient settings.
Especially, using only 50\% labelled training data, it is able to achieve comparable performance with the baseline using 100\% labelled data.

\section*{Acknowledgement}
\noindent This research is supported by the Agency for Science, Technology and Research (A*STAR), Singapore under its MTC Young Individual Research Grant (Grant No. M21K3c0130) and Career Development Award (Grant No. 202D8243).

{\small
\bibliographystyle{ieee_fullname}
\bibliography{pc}

\begin{thebibliography}{10}\itemsep=-1pt

\bibitem{lgan18}
Panos Achlioptas, Olga Diamanti, Ioannis Mitliagkas, and Leonidas~J. Guibas.
\newblock Learning representations and generative models for 3d point clouds.
\newblock In {\em ICML}, volume~80, pages 40--49. {PMLR}, 2018.

\bibitem{s3dis}
Iro Armeni, Ozan Sener, Amir~Roshan Zamir, Helen Jiang, Ioannis~K. Brilakis,
  Martin Fischer, and Silvio Savarese.
\newblock 3d semantic parsing of large-scale indoor spaces.
\newblock In {\em CVPR}, pages 1534--1543, 2016.

\bibitem{shapegf20}
Ruojin Cai, Guandao Yang, Hadar Averbuch-Elor, Zekun Hao, Serge Belongie, Noah
  Snavely, and Bharath Hariharan.
\newblock Learning gradient fields for shape generation.
\newblock In {\em ECCV}, 2020.

\bibitem{simclr20}
Ting Chen, Simon Kornblith, Mohammad Norouzi, and Geoffrey~E. Hinton.
\newblock A simple framework for contrastive learning of visual
  representations.
\newblock {\em CoRR}, abs/2002.05709, 2020.

\bibitem{semi21_aaai}
Mingmei Cheng, Le Hui, Jin Xie, and Jian Yang.
\newblock Sspc-net: Semi-supervised semantic 3d point cloud segmentation
  network.
\newblock In {\em AAAI}, pages 1140--1147. {AAAI} Press, 2021.

\bibitem{svm}
C. Cortes and V. Vapnik.
\newblock Support vector networks.
\newblock {\em Machine Learning}, 20, 1995.

\bibitem{cbn17}
Harm de Vries, Florian Strub, J{\'{e}}r{\'{e}}mie Mary, Hugo Larochelle,
  Olivier Pietquin, and Aaron~C. Courville.
\newblock Modulating early visual processing by language.
\newblock In {\em NIPS}, pages 6594--6604, 2017.

\bibitem{discrim14}
Alexey Dosovitskiy, Jost~Tobias Springenberg, Martin~A. Riedmiller, and Thomas
  Brox.
\newblock Discriminative unsupervised feature learning with convolutional
  neural networks.
\newblock In {\em NIPS}, pages 766--774, 2014.

\bibitem{ParAE21}
Benjamin Eckart, Wentao Yuan, Chao Liu, and Jan Kautz.
\newblock Self-supervised learning on 3d point clouds by learning discrete
  generative models.
\newblock In {\em CVPR}, pages 8248--8257, 2021.

\bibitem{ACD20}
Matheus Gadelha, Aruni RoyChowdhury, Gopal Sharma, Evangelos Kalogerakis,
  Liangliang Cao, Erik~G. Learned{-}Miller, Rui Wang, and Subhransu Maji.
\newblock Label-efficient learning on point clouds using approximate convex
  decompositions.
\newblock In {\em ECCV}, volume 12355, pages 473--491, 2020.

\bibitem{mrtnet18}
Matheus Gadelha, Rui Wang, and Subhransu Maji.
\newblock Multiresolution tree networks for 3d point cloud processing.
\newblock In {\em ECCV}, volume 11211 of {\em Lecture Notes in Computer
  Science}, pages 105--122, 2018.

\bibitem{rot18}
Spyros Gidaris, Praveer Singh, and Nikos Komodakis.
\newblock Unsupervised representation learning by predicting image rotations.
\newblock In {\em ICLR}, 2018.

\bibitem{tlnet16}
Rohit Girdhar, David~F. Fouhey, Mikel Rodriguez, and Abhinav Gupta.
\newblock Learning a predictable and generative vector representation for
  objects.
\newblock In {\em ECCV}, volume 9910, pages 484--499, 2016.

\bibitem{gan}
Ian~J. Goodfellow, Jean Pouget{-}Abadie, Mehdi Mirza, Bing Xu, David
  Warde{-}Farley, Sherjil Ozair, Aaron~C. Courville, and Yoshua Bengio.
\newblock Generative adversarial nets.
\newblock In {\em NIPS}, pages 2672--2680, 2014.

\bibitem{contrastive06}
Raia Hadsell, Sumit Chopra, and Yann LeCun.
\newblock Dimensionality reduction by learning an invariant mapping.
\newblock In {\em CVPR}, pages 1735--1742, 2006.

\bibitem{vipgan19}
Zhizhong Han, Mingyang Shang, Yu{-}Shen Liu, and Matthias Zwicker.
\newblock View inter-prediction {GAN:} unsupervised representation learning for
  3d shapes by learning global shape memories to support local view
  predictions.
\newblock In {\em AAAI}, pages 8376--8384. {AAAI} Press, 2019.

\bibitem{mapvae19}
Zhizhong Han, Xiyang Wang, Yu{-}Shen Liu, and Matthias Zwicker.
\newblock Multi-angle point cloud-vae: Unsupervised feature learning for 3d
  point clouds from multiple angles by joint self-reconstruction and
  half-to-half prediction.
\newblock In {\em ICCV}, 2019.

\bibitem{multitask19}
Kaveh Hassani and Mike Haley.
\newblock Unsupervised multi-task feature learning on point clouds.
\newblock In {\em ICCV}, pages 8159--8170, 2019.

\bibitem{moco20}
Kaiming He, Haoqi Fan, Yuxin Wu, Saining Xie, and Ross~B. Girshick.
\newblock Momentum contrast for unsupervised visual representation learning.
\newblock In {\em CVPR}, pages 9726--9735, 2020.

\bibitem{resnet}
Kaiming He, Xiangyu Zhang, Shaoqing Ren, and Jian Sun.
\newblock Deep residual learning for image recognition.
\newblock In {\em CVPR}, pages 770--778, 2016.

\bibitem{mutualinfo19}
R.~Devon Hjelm, Alex Fedorov, Samuel Lavoie{-}Marchildon, Karan Grewal, Philip
  Bachman, Adam Trischler, and Yoshua Bengio.
\newblock Learning deep representations by mutual information estimation and
  maximization.
\newblock In {\em ICLR}, 2019.

\bibitem{Huang21}
Siyuan Huang, Yichen Xie, Song{-}Chun Zhu, and Yixin Zhu.
\newblock Spatio-temporal self-supervised representation learning for 3d point
  clouds.
\newblock In {\em ICCV}, pages 6515--6525. {IEEE}, 2021.

\bibitem{semi21}
Li Jiang, Shaoshuai Shi, Zhuotao Tian, Xin Lai, Shu Liu, Chi{-}Wing Fu, and
  Jiaya Jia.
\newblock Guided point contrastive learning for semi-supervised point cloud
  semantic segmentation.
\newblock 2021.

\bibitem{gcn17}
Thomas~N. Kipf and Max Welling.
\newblock Semi-supervised classification with graph convolutional networks.
\newblock In {\em ICLR}. OpenReview.net, 2017.

\bibitem{sonet18}
Jiaxin Li, Ben~M. Chen, and Gim~Hee Lee.
\newblock So-net: Self-organizing network for point cloud analysis.
\newblock In {\em CVPR}, pages 9397--9406, 2018.

\bibitem{vsl18}
Shikun Liu, C.~Lee Giles, and Alexander Ororbia.
\newblock Learning a hierarchical latent-variable model of 3d shapes.
\newblock In {\em ICDV}, pages 542--551, 2018.

\bibitem{occnet19}
Lars~M. Mescheder, Michael Oechsle, Michael Niemeyer, Sebastian Nowozin, and
  Andreas Geiger.
\newblock Occupancy networks: Learning 3d reconstruction in function space.
\newblock In {\em CVPR}, pages 4460--4470, 2019.

\bibitem{jigsaw16}
Mehdi Noroozi and Paolo Favaro.
\newblock Unsupervised learning of visual representations by solving jigsaw
  puzzles.
\newblock In {\em ECCV}, volume 9910, pages 69--84, 2016.

\bibitem{occo2d}
Deepak Pathak, Philipp Kr{\"{a}}henb{\"{u}}hl, Jeff Donahue, Trevor Darrell,
  and Alexei~A. Efros.
\newblock Context encoders: Feature learning by inpainting.
\newblock In {\em CVPR}, pages 2536--2544, 2016.

\bibitem{rotation3d}
Omid Poursaeed, Tianxing Jiang, Han Qiao, Nayun Xu, and Vladimir~G. Kim.
\newblock Self-supervised learning of point clouds via orientation estimation.
\newblock In Vitomir Struc and Francisco~G{\'{o}}mez Fern{\'{a}}ndez, editors,
  {\em 3DV}, pages 1018--1028, 2020.

\bibitem{pointnet2}
Charles~Ruizhongtai Qi, Li Yi, Hao Su, and Leonidas~J. Guibas.
\newblock Pointnet++: Deep hierarchical feature learning on point sets in a
  metric space.
\newblock In {\em NIPS}, pages 5099--5108, 2017.

\bibitem{scannet10}
Can Qin, Haoxuan You, Lichen Wang, C.{-}C.~Jay Kuo, and Yun Fu.
\newblock Pointdan: {A} multi-scale 3d domain adaption network for point cloud
  representation.
\newblock In Hanna~M. Wallach, Hugo Larochelle, Alina Beygelzimer, Florence
  d'Alch{\'{e}}{-}Buc, Emily~B. Fox, and Roman Garnett, editors, {\em NIPS},
  pages 7190--7201, 2019.

\bibitem{glr20}
Yongming Rao, Jiwen Lu, and Jie Zhou.
\newblock Global-local bidirectional reasoning for unsupervised representation
  learning of 3d point clouds.
\newblock In {\em CVPR}, pages 5375--5384, 2020.

\bibitem{selfsup19}
Jonathan Sauder and Bjarne Sievers.
\newblock Self-supervised deep learning on point clouds by reconstructing
  space.
\newblock In {\em NIPS}, pages 12942--12952, 2019.

\bibitem{Shi20}
Yi Shi, Mengchen Xu, Shuaihang Yuan, and Yi Fang.
\newblock Unsupervised deep shape descriptor with point distribution learning.
\newblock In {\em CVPR}, pages 9350--9359, 2020.

\bibitem{Sitzmann19}
Vincent Sitzmann, Michael Zollh{\"{o}}fer, and Gordon Wetzstein.
\newblock Scene representation networks: Continuous 3d-structure-aware neural
  scene representations.
\newblock In {\em NIPS}, pages 1119--1130, 2019.

\bibitem{scanobjnn}
Mikaela~Angelina Uy, Quang{-}Hieu Pham, Binh{-}Son Hua, Duc~Thanh Nguyen, and
  Sai{-}Kit Yeung.
\newblock Revisiting point cloud classification: {A} new benchmark dataset and
  classification model on real-world data.
\newblock In {\em ICCV}, pages 1588--1597, 2019.

\bibitem{infonce18}
A{\"{a}}ron van~den Oord, Yazhe Li, and Oriol Vinyals.
\newblock Representation learning with contrastive predictive coding.
\newblock {\em CoRR}, abs/1807.03748, 2018.

\bibitem{tsne}
L.J.P. Van~der Maaten and G.E. Hinton.
\newblock Visualizing high-dimensional data using t-sne.
\newblock 2008.

\bibitem{occnet21}
Hanchen Wang, Qi Liu, Xiangyu Yue, Joan Lasenby, and Matt~J. Kusner.
\newblock Unsupervised point cloud pre-training via occlusion completion.
\newblock In {\em ICCV}, 2021.

\bibitem{Wei20}
Jiacheng Wei, Guosheng Lin, Kim{-}Hui Yap, Tzu{-}Yi Hung, and Lihua Xie.
\newblock Multi-path region mining for weakly supervised 3d semantic
  segmentation on point clouds.
\newblock In {\em CVPR}, pages 4383--4392. Computer Vision Foundation / {IEEE},
  2020.

\bibitem{3dgan16}
Jiajun Wu, Chengkai Zhang, Tianfan Xue, Bill Freeman, and Josh Tenenbaum.
\newblock Learning a probabilistic latent space of object shapes via 3d
  generative-adversarial modeling.
\newblock In {\em NIPS}, pages 82--90, 2016.

\bibitem{modelnet}
Zhirong Wu, Shuran Song, Aditya Khosla, Fisher Yu, Linguang Zhang, Xiaoou Tang,
  and Jianxiong Xiao.
\newblock 3d shapenets: {A} deep representation for volumetric shapes.
\newblock In {\em CVPR}, pages 1912--1920. {IEEE} Computer Society, 2015.

\bibitem{pointcontrast20}
Saining Xie, Jiatao Gu, Demi Guo, Charles~R Qi, Leonidas~J Guibas, and Or
  Litany.
\newblock Pointcontrast: Unsupervised pre-training for 3d point cloud
  understanding.
\newblock 2020.

\bibitem{Xu20}
Xun Xu and Gim~Hee Lee.
\newblock Weakly supervised semantic point cloud segmentation: Towards 10x
  fewer labels.
\newblock In {\em CVPR}, pages 13703--13712. Computer Vision Foundation /
  {IEEE}, 2020.

\bibitem{foldingnet18}
Yaoqing Yang, Chen Feng, Yiru Shen, and Dong Tian.
\newblock Foldingnet: Point cloud auto-encoder via deep grid deformation.
\newblock In {\em CVPR}, pages 206--215. {IEEE} Computer Society, 2018.

\bibitem{shapenetpart}
Li Yi, Vladimir~G. Kim, Duygu Ceylan, I{-}Chao Shen, Mengyan Yan, Hao Su, Cewu
  Lu, Qixing Huang, Alla Sheffer, and Leonidas~J. Guibas.
\newblock A scalable active framework for region annotation in 3d shape
  collections.
\newblock {\em {ACM} Trans. Graph.}, 35(6):210:1--210:12, 2016.

\bibitem{clusternet19}
Ling Zhang and Zhigang Zhu.
\newblock Unsupervised feature learning for point cloud understanding by
  contrasting and clustering using graph convolutional neural networks.
\newblock In {\em ICDV}, pages 395--404, 2019.

\bibitem{color16}
Richard Zhang, Phillip Isola, and Alexei~A. Efros.
\newblock Colorful image colorization.
\newblock In Bastian Leibe, Jiri Matas, Nicu Sebe, and Max Welling, editors,
  {\em ECCV}, volume 9907, pages 649--666, 2016.

\bibitem{depthcontrast21}
Zaiwei Zhang, Rohit Girdhar, Armand Joulin, and Ishan Misra.
\newblock Self-supervised pretraining of 3d features on any point-cloud.
\newblock 2021.

\bibitem{3dcaps19}
Yongheng Zhao, Tolga Birdal, Haowen Deng, and Federico Tombari.
\newblock 3d point capsule networks.
\newblock In {\em CVPR}, pages 1009--1018, 2019.

\end{thebibliography}
}

\end{document}